%% file: main.tex
\documentclass[a4paper,conference]{IEEEtran}
\IEEEoverridecommandlockouts
\usepackage{cite}
\usepackage{multirow}
\ifCLASSINFOpdf
\usepackage[pdftex]{graphicx}
\else
\fi
\begin{document}
\newcommand{\etal}{\textit{et al}.}

%
% paper title
% Titles are generally capitalized except for words such as a, an, and, as,
% at, but, by, for, in, nor, of, on, or, the, to and up, which are usually
% not capitalized unless they are the first or last word of the title.
% Linebreaks \\ can be used within to get better formatting as desired.
% Do not put math or special symbols in the title.
\title{Adaptive Instance Distillation\\ for Object Detection in Autonomous Driving}

% author names and affiliations
% use a multiple column layout for up to three different
% affiliations
\author{\IEEEauthorblockN{Qizhen Lan and Qing Tian\IEEEauthorrefmark{2}\thanks{\IEEEauthorrefmark{2}Corresponding Author.}}
\IEEEauthorblockA{
Department of Computer Science, Bowling Green State University\\
Bowling Green, Ohio 43403, USA\\
Email: \{qlan,qtian\}@bgsu.edu}
%\and
%\IEEEauthorblockN{Qing Tian}
%\IEEEauthorblockA{Department of Computer Science, Bowling Green State University\\
%Bowling Green, Ohio 43403\\
%Email: qtian@bgsu.edu}
}

% conference papers do not typically use \thanks and this command
% is locked out in conference mode. If really needed, such as for
% the acknowledgment of grants, issue a \IEEEoverridecommandlockouts
% after \documentclass

% for over three affiliations, or if they all won't fit within the width
% of the page, use this alternative format:
%
%\author{\IEEEauthorblockN{Michael Shell\IEEEauthorrefmark{1},
%Homer Simpson\IEEEauthorrefmark{2},
%James Kirk\IEEEauthorrefmark{3},
%Montgomery Scott\IEEEauthorrefmark{3} and
%Eldon Tyrell\IEEEauthorrefmark{4}}
%\IEEEauthorblockA{\IEEEauthorrefmark{1}School of Electrical and Computer Engineering\\
%Georgia Institute of Technology,
%Atlanta, Georgia 30332--0250\\ Email: see http://www.michaelshell.org/contact.html}
%\IEEEauthorblockA{\IEEEauthorrefmark{2}Twentieth Century Fox, Springfield, USA\\
%Email: homer@thesimpsons.com}
%\IEEEauthorblockA{\IEEEauthorrefmark{3}Starfleet Academy, San Francisco, California 96678-2391\\
%Telephone: (800) 555--1212, Fax: (888) 555--1212}
%\IEEEauthorblockA{\IEEEauthorrefmark{4}Tyrell Inc., 123 Replicant Street, Los Angeles, California 90210--4321}}

% use for special paper notices
%\IEEEspecialpapernotice{(Invited Paper)}

% make the title area
\maketitle

% As a general rule, do not put math, special symbols or citations
% in the abstract
\begin{abstract}
In recent years, knowledge distillation (KD) has been widely used to derive efficient models. Through imitating a large teacher model, a lightweight student model can achieve comparable performance with more efficiency. However, most existing knowledge distillation methods are focused on classification tasks. Only a limited number of studies have applied knowledge distillation to object detection, especially in time-sensitive autonomous driving scenarios. In this paper, we propose Adaptive Instance Distillation (AID) to selectively impart teacher's knowledge to the student to improve the performance of knowledge distillation. Unlike previous KD methods that treat all instances equally, our AID can attentively adjust the distillation weights of instances based on the teacher model's prediction loss. We verified the effectiveness of our AID method through experiments on the KITTI and the COCO traffic datasets. The results show that our method improves the performance of state-of-the-art attention-guided and non-local distillation methods and achieves better distillation results on both single-stage and two-stage detectors. Compared to the baseline, our AID led to an average of 2.7\% and 2.1\% mAP increases for single-stage and two-stage detectors, respectively. Furthermore, our AID is also shown to be useful for self-distillation to improve the teacher model's performance.
\end{abstract}

% no keywords
% For peer review papers, you can put extra information on the cover
% page as needed:
% \ifCLASSOPTIONpeerreview
% \begin{center} \bfseries EDICS Category: 3-BBND \end{center}
% \fi
%
% For peerreview papers, this IEEEtran command inserts a page break and
% creates the second title. It will be ignored for other modes.
\IEEEpeerreviewmaketitle

\section{Introduction}
% no \IEEEPARstart

Knowledge Distillation (KD) \cite{hinton2015distilling, krizhevsky2009learning} has been introduced to derive a high-performance and lightweight student model by mimicking the knowledge from a large, powerful, and computationally intensive teacher model. Many KD methods have been explored so far and have achieved promising results in classification problems. However, only a limited number of studies have applied KD to the more challenging object detection task. Most KD in object detection methods investigate what types of knowledge should be mimicked, e.g., feature maps \cite{DBLP:journals/corr/abs-1906-03609}, head soft prediction \cite{DBLP:journals/corr/abs-2102-12252}, attention-based feature map \cite{zhang2021improve} or relation between instances \cite{DBLP:journals/corr/abs-2103-02340}. They usually treat all instances equally when transferring the location and category knowledge from the teacher to the student. However, teachers do not learn the instances equally well. We argue that the distillation process should adaptively change focus based on different instances.
In other words, knowledge distillation should not only pay attention to what kind of knowledge to imitate, but also to which instances are more valuable.
Specifically, knowledge from instances that the teacher can accurately predict should be recognized and transferred to the student, and the student should avoid paying too much attention to the ``unsure" instances from the teacher's perspective.

Instance reweighting has been used to improve models' performance in object detection by assigning larger weights to important instances \cite{FL, li2018gradient, cao2019prime}. However, Zhang \etal~\cite{zhang2020primeaware} have shown that the hard mining methods for object detection are not suitable for knowledge distillation.
Zhang \etal~\cite{zhang2020primeaware} added an auxiliary task branch to the student model, and the variance of the features from that auxiliary branch, which they called data uncertainty, is utilized for distillation reweighting.

In this paper, we propose a new method called Adaptive Instance Distillation (AID) that reweights distilled instances based on teacher-judged difficulty. In contrast to Zhang \etal~\cite{zhang2020primeaware}, our AID does not employ an uncertainty estimation method based on auxiliary tasks because the variance of auxiliary features may not always represent the task uncertainty and it results in additional computation. More importantly, we argue that the importance of instances should not be determined by the feature statistics of the student network but rather by the teacher's prediction. Our AID reweights an instance based on the teacher's original loss, which reflects the reliability of the teacher on that instance. Specifically, an instance with larger teacher's prediction losses will receive small distillation weights and thus less attention from the student model. In other words, our AID allows the student to learn more from the teacher on instances that the teacher performs well, while giving the student more freedom to learn ``teacher-uncertain" instances on their own. It is worth mentioning that most state-of-the-art object detectors adopt Feature Pyramid Networks (FPN) \cite{fpn} in their detection pipeline.
Our AID can be applied to each scale of the FPN output with the potential to improve the detection ability of multi-scale objects. 

In summary, this paper has the following contributions:
\begin{enumerate}
    \item For the first time in KD, we propose that students should selectively learn from their teachers on a per-instance basis according to teacher's performance rather than blindly learning from all instances equally.
    \item We propose Adaptive Instance Distillation (AID) to allow students to discern the reliability of the teacher's knowledge regarding a particular instance based on the teacher's predictions losses.
    \item Our AID re-weighing method can be applied to each layer of FPN to achieve scale-wise selection in knowledge distillation.
    \item Our AID is suitable for both single-stage and two-stage detectors, and it can improve the distilled model's ability to detect difficult instances compared to several baseline networks.
\end{enumerate}

% You must have at least 2 lines in the paragraph with the drop letter
% (should never be an issue)

\section{Related works}

This section reviews the most relevant works in the areas of Object Detection, Adaptive Sample Weighting, and Knowledge Distillation.

\paragraph{Object Detection}
Some pioneers of object detectors include \cite{Vaillant94anoriginal, 990517}. Nowadays, object detection models are generally classified into two categories: \textbf{Two-stage Detectors}. Typical examples include Faster R-CNN \cite{DBLP:journals/corr/RenHG015}, where in the first stage, the Region Proposal Network (RPN) is employed to generate a set of proposals for potential objects, then in the second stage, classification and localization are made on selected proposed regions. The other type of detector is \textbf{One-stage Detectors}. Some examples include \cite{YOLOV1, FL, li2020generalized}. They are well known for their high efficiency compared to two-stage detectors. They perform classification and localization prediction directly without proposals for regions of interest. Recently, single-stage detectors have been further divided into anchor-based and anchor-free detectors. Anchor-based detectors, such as \cite{DBLP:journals/corr/RedmonF16, FL}, need to traverse a large number of anchor boxes to ensure the accuracy of the entire detection task. The anchor-free methods \cite{yolox, FCOS} directly predict the center point or key-points of an object from feature maps, which reduces the computational cost.
The well-known method FPN \cite{fpn} or its variant \cite{DBLP:journals/corr/abs-2006-02334} has been adopted by many state-of-the-art detectors to improve the ability to detect objects of different scales.

\paragraph{Adaptive Instance Weighting}

Adaptive instance weighting by adjusting the contribution of each instance can help with effective learning in object detection. ``Hard example mining" is one reweighting technique that puts non-uniform attention to samples based on difficulty. In object detection works \cite{li2018gradient, FL, cao2019prime}, hard-mining plays a critical role in improving model detection performance. Typically, hard mining gives larger weights to hard instances and forces the model to pay more attention to the complex instances during training. However, Zhang \etal~\cite{zhang2020primeaware} argue that hard-mining may not be appropriate for knowledge distillation. In contrast, down-weighting those hard samples or paying more attention to easy samples will allow the distillation model to achieve better performance. One important question to ask is: how should the example difficulty be measured? In object detection, He \etal~\cite{FL} use modified cross-entropy loss (a.k.a. focal loss) to measure the difficulty of bounding boxes. Bounding boxes with high prediction probability of the correct class (e.g., most backgrounds) are considered to be easy and they receive even less attention compared to the unmodified cross entropy case. On the other hand, the modified loss directs more attention to the not-so-well-classified examples. GHM-C in \cite{li2018gradient} follows a similar idea to focal loss.
Cao \etal~\cite{cao2019prime} use Hierarchical Local Ranks to compute sample importance in mini-batches. 
In knowledge distillation, Zhang \etal~\cite{zhang2020primeaware} measure instance importance through feature variance of an auxiliary branch added to the student model.
As in~\cite{zhang2020primeaware}, our method applies instance reweighting during knowledge distillation. However, unlike previous approaches, we utilize the teacher network's predictions to determine instance importance for the student.

\paragraph{Knowledge Distillation}

Since its introduction by Hinton \etal~\cite{hinton2015distilling}, knowledge distillation (KD) has been widely applied in deep learning. The goal of KD is to train a new high-performance lighter-weight student network by transferring a powerful teacher network's knowledge. In knowledge distillation, the knowledge comes in different forms: feature-based \cite{Romero15fitnets:hints, huang2017like, VID2019, Byeongho2018, 8100237, zhang2021improve}, response-based \cite{hinton2015distilling, DBLP:journals/corr/abs-2102-12252} and relation-based \cite{8953802, DBLP:journals/corr/abs-1904-05068, DBLP:journals/corr/abs-2103-02340}. The main difference lies in the kind of the knowledge transferred. Unlike the distillation for classification problems, improving the efficiency and performance of an object detection model through knowledge distillation is a more challenging task. Relatively speaking, KD is less explored in object detection than in classification tasks. It was not until 2017 that Chen \etal \space \cite{Chen2017LearningEO} first proposed their KD method for object detection. To deal with the high imbalance between the background and foreground object regions of interest in object detection, Chen \etal \space \cite{Chen2017LearningEO} down-weighted the background distillation loss in the classification head. Zhang \etal~\cite{zhang2021improve} proposed to use an attention-based method to improve the distillation results. However, to the best of our knowledge, there is only one work \cite{zhang2020primeaware} that attempts to apply the idea of instance-based reweighting to the domain of distillation. They used an auxiliary task branch to flatten the student's feature map into a vector and that vector's variance is utilized to adjust the distillation weights. They give larger weights to samples with low variances.
However, there is no enough justification why auxiliary feature variance and sample importance are related.
In contrast to Zhang \etal~\cite{zhang2020primeaware} that uses the student network's information to measure instance weight, we leverage the teacher's prediction for each instance to determine the reliability of the distilled knowledge.

\section{Methodology}

Object detection involves multiple tasks, e.g., bounding box regression, category classification, and objectness prediction. Therefore, knowledge distillation for object detection is more complex than for classification. In an image to be detected, the background instances are often more pervasive than the foreground object instances. To deal with the imbalance problem, many adaptive instance weighting strategies, such as hard sample mining \cite{FL}, have been proposed.
However, Zhang \etal \space \cite{zhang2020primeaware} show that hard sample mining does not work well in knowledge distillation. Instead, they used an auxiliary task branch to estimate uncertainty in the data and make students pay more attention to the `stable' samples. However, the variety of the auxiliary features is not necessarily a reliable indicator for instance importance, and it does not represent the importance of the knowledge from the teacher. In contrast to their approach, we propose to measure the value of the teacher's knowledge on a per-instance basis by calculating the gap between the ground truth and the teacher's prediction. In other words, if the teacher model cannot predict an example well, it implies that the teacher's knowledge about that instance is less trustworthy. On the other hand, valuable knowledge comes from those instances that can be accurately predicted by the teacher model. The student network should pay more attention to such instances.

\subsection{Adaptive Instance Knowledge Distillation}
\input{fig/AID}

In general, knowledge distillation tasks have two kinds of losses. One is the distillation loss $\mathcal{L}_{distill}$ which measures the knowledge (or prediction) difference between the student and the teacher model. The other one is the task loss, which is used to guide the student to learn the original task. 
In this paper, we propose Adaptive Instance distillation (AID) to adaptively distill the knowledge of the teacher model on a per-instance basis for object detection tasks. The idea is that the student model should pay more attention to instances in which the teacher has more authority/trustworthiness rather than learn all instances equally from the teacher model. 
Fig. \ref{fig:AID} illustrates how AID guides the student model to better learn the most valuable and reliable knowledge from the teacher model.
We define the overall loss for student learning as:

\begin{equation}
    \mathcal{L}_{i}^{\mathcal{S}} = \mathcal{L}_{task,i}^{\mathcal{S}} + \lambda\mathcal{L}^{\mathcal{S,T}}_{AID,i} \,.
\end{equation}

\noindent where $i$ indicates the $i$-th instance. The superscripts $\mathcal{S}$ and $\mathcal{T}$ imply that a corresponding loss term depends on the student and/or the teacher prediction.
$\lambda$ is a weighting factor balancing the contribution between the task loss $\mathcal{L}_{task}$ and our instance adaptive distillation loss $\mathcal{L}_{AID}$. The latter is defined as follows:

\begin{equation}\label{eq:negexp}
    \mathcal{L}^{\mathcal{S,T}}_{AID,i} = \exp^{-\alpha\mathcal{D}_{i}^\mathcal{T}}\mathcal{L}^{\mathcal{S,T}}_{distill,i} \,,
\end{equation}

\noindent where

\begin{equation}
    \mathcal{D}_{i}^\mathcal{T} = \mathcal{L}_{task,i}^\mathcal{T} %\,,
\end{equation}

\noindent is the teacher's object detection task loss, i.e., the distance between the ground truth and the prediction, on the $i$-th instance. $\alpha$ is a hyper-parameter that needs to be tuned empirically (we set it to 0.1 in all our experiments).
As we can see from Eq.~\ref{eq:negexp}, the adaptive weight of the instance $i$ has a negative exponential correlation with the teacher's prediction loss. 
The larger the teacher's error on a certain instance $i$ (i.e., $\mathcal{D}_i^\mathcal{T}$) is, the smaller weight or less attention the instance $i$ will receive from the student model during the knowledge distillation process. On the other hand, instances where the teacher predicts accurately (i.e., with smaller $\mathcal{D}_{i}^T$ values) deserve more of the student's attention in the knowledge transfer process.
The instance weight degrades exponentially with the increase of the teacher's prediction error. The exponential function sets an appropriate range of the punishment. Take the extreme cases for example. An instance where the teacher's loss is extremely large will receive approximately zero attention while there will be no knowledge transfer degradation for instances where the teacher model makes `perfect' prediction (zero task loss).  

Putting all things together, we get the final loss for our instance-adaptive student learning:

\begin{equation}\label{eq:Usual KD loss}
    \mathcal{L}_{i}^\mathcal{S} = \mathcal{L}^\mathcal{S}_{task,i} + \lambda\exp^{-\alpha\mathcal{L}_{task,i}^\mathcal{T}}\mathcal{L}_{distill,i}^{\mathcal{S,T}} \,.
\end{equation}

Feature Pyramid Networks (FPN) \cite{fpn} have been widely adopted by state-of-the-art object detectors. To improve knowledge transfer for objects of different scales, we apply our AID strategy to each output layer of the FPN. 
In this case, our AID adaptively weighs not only the instance-wise knowledge but also scale-wise feature knowledge during the knowledge distillation process. 
The student will rely more on the teacher for scales where the teacher feels more confident\footnote{here, confident is loosely defined as knowledgeable}. For scales where the teacher performs badly, the student will rely on itself to learn instead of being misled by the teacher. 
Such scale-adaptive knowledge distillation contributes to better object detection on different scales. More details will follow in the experiment section.

\subsection{Instance-adaptive Self Knowledge Distillation}

Although some effective self-knowledge-distillation approaches \cite{DBLP:journals/corr/abs-1905-08094} exist for classification tasks, relatively few works have explored self-distillation on detection models.
In addition to help better distill knowledge from a teacher model to a more compact student model, our AID strategy can also be used to distill useful knowledge into the original architecture itself. In this adaptive self-knowledge-distillation process, the knowledge from easy instances will be passed on to the new network with the same architecture, without much information loss. On the other hand, we pay less attention to or even discard the knowledge gained from difficult instances where the original model does not perform well. This provides an opportunity for the new network to re-learn those difficult examples given the knowledge from some easy instances.
This instance-adaptive self-reflection process is expected to result in an improved detection model with the same architecture.
Our AID-based self-distillation loss for object detection is defined as follows:

\begin{equation}
    \mathcal{L}^{new}_{i} = \mathcal{L}^{new}_{task,i} + \lambda\exp^{-\alpha\mathcal{L}_{task,i}^{old}} \mathcal{L}^{new,old}_{distill,i} \,,
\end{equation}

\noindent where $new$ represents the new model we want to train, $old$ indicates the previous pre-trained model. In our experiments, we conducted self-distillation experiments on both single-stage and two-stage detectors.

\section{Experiments and Results}
\subsection{Experimental Setup}

\subsubsection{Datasets}
To evaluate our method, we utilize two autonomous driving related datasets in our experiments. The first one is the KITTI 2D-object detection dataset \cite{Geiger2012CVPR}, which includes three different types of road objects (i.e., car, cyclist, and pedestrian). The second one is the COCO traffic dataset, which is obtained by selecting categories related to self-driving from MS COCO 2017 \cite{lin2015microsoft}. These categories include: \textit{person, stop sign, traffic light, fire hydrant, parking meter, bus, motorcycle, bicycle, car, train, truck}.

\subsubsection{Implementation Details}

All the detection methods, including the baselines, are implemented within the MMDetection \cite{mmdetection} framework. In our experiments, we chose Faster-RCNN \cite{DBLP:journals/corr/RenHG015} as an example of two-stage detectors and Generalized Focal Loss (GFL) \cite{li2020generalized} as an example of single-stage detectors. We experimented with two backbone architectures (i.e., ResNet101 and ResNet-50) on both the single-stage and two-stage detectors. The teacher model was directly trained with MMDetection's default configuration (without any KD methods).
We adopted one state-of-the-art KD method (Zhang \etal~\cite{zhang2021improve}) as the KD baseline and applied AID to verify our AID's effectiveness. On the two datasets, we used the same experimental settings and hyperparameters. We do not perform much hyperparameter tweaking, and the default hyperparameters in the pre-configured file are adopted.
In our implementation, the hyperparameters of Zhang \etal~\cite{zhang2021improve}'s method are set as $\{\beta = 4 \times 10^{-3}, \gamma = \eta = 7 \times 10^{-5}, T = 0.5\}$ for Faster R-CNN, $\{\beta = 2 \times 10^{-2}, \gamma = \eta = 4 \times 10^{-4}, T=0.1\}$ for GFL. We set $\alpha=0.1$ for all knowledge distillation. All models are sufficiently trained to convergence (i.e., 24 epochs for models with ResNet101 backbone, 12 epochs for ResNet50 backbone). All models are evaluated in terms of mean averaged precision (mAP) with 0.5 as the Intersection over Union (IoU) threshold.

\subsection{Quantitative Analysis}

We compare our Adaptive Instance Distillation (AID) method with the baseline and one state-of-the-art KD method. The baseline is trained with a certain backbone without any knowledge distillation. The compared KD method is an attention-guided knowledge distillation method of Zhang \etal \space \cite{zhang2021improve}. 
We evaluate our AID method by applying it to Zhang \etal \space \cite{zhang2021improve} to see whether our AID can improve the distillation results.
In our experiments, all teacher models employ a ResNet-101 backbone. In addition to using ResNet-50 backbones for the students, we also employ students with ResNet-101 backbones to perform self-distillation and evaluate our AID-based self-distillation's effectiveness.

\begin{table*}[h!]
\footnotesize
\caption{Results on KITTI dataset. Student-Baseline and Teacher-Baseline refer to non-distillation models with ResNet-50 and ResNet-101 backbones, respectively. Zhang \etal\cite{zhang2021improve}: a state-of-the-art distillation baseline without adaptive weighting.}
\centering
\begin{tabular}{c|c|c|c|c}
\hline
Teacher-Backbone & Student-Backbone & Method & mAP (GFL) & mAP (Faster R-CNN)\\
\hline
\multirow{6}{*}{ResNet-101} & \multirow{3}{*}{ResNet-50}  & Student-Baseline &85.1 & 88.9 \\
 & & Zhang et al. \cite{zhang2021improve} &86.4  & 89.0 \\ 
 & & Ours (AID) &\textbf{88.0}  & \textbf{89.6} \\\cline{2-5}
 & \multirow{3}{*}{ResNet-101} & Teacher-Baseline & 89.4  & 89.6 \\
 & & Zhang et al. \cite{zhang2021improve}&88.2  & 89.7  \\
 & & Ours (AID) &\textbf{89.7}  & \textbf{89.8} \\
\hline
\end{tabular}
\label{tab:KITTI}
\end{table*}

Table \ref{tab:KITTI} shows the comparison between our method and the baselines in different scenarios (e.g., different student backbones, one-stage vs. two stages detectors). It can be observed that the attention-based KD method (Zhang \etal~\cite{zhang2021improve}) can improve performance to a limited extent (for example, the mAP of the single-stage detector is increased by 1.3\%). After we applied our method, those improvements became more significant (mAP got increased by 2.9\%). In the self-distillation scenario, our approach can also improve the teacher model's performance by 0.5\% without any architecture changes or extra teachers.

Our AID-based approach achieves promising results on the COCO traffic dataset as well (Table \ref{tab:COCO}). When using ResNet-50 as the student backbone, our AID method achieves 2.5\% and 0.7\% mAP improvement over the teacher baseline for the GFL and Faster R-CNN cases, respectively.
In self-distillation scenarios, our AID approach also beats all other baselines. Compared to the teacher baseline (also with a ResNet-101 backbone), our AID achieves 1.6\% mAP improvement in the GFL case and 1.0\% in the Faster R-CNN case.

\begin{table*}[h!]
\footnotesize
\caption{Results on COCO traffic dataset. Student-Baseline and Teacher-Baseline refer to non-distillation models with ResNet-50 and ResNet-101 backbones, respectively. Zhang \etal\cite{zhang2021improve}: a state-of-the-art distillation baseline without adaptive weighting.}
\centering
\begin{tabular}{c|c|c|c|c}
\hline
Teacher-Backbone & Student-Backbone & Method & mAP (GFL)  & mAP (Faster R-CNN) \\
\hline
\multirow{6}{*}{ResNet-101} & \multirow{3}{*}{ResNet-50}  & Student-Baseline & 67.6  & 65.3 \\
 & & Zhang et al. \cite{zhang2021improve} & 69.7  & 67.9  \\ 
 & & Ours (AID) & \textbf{70.1}  &\textbf{68.7} \\\cline{2-5}
 & \multirow{3}{*}{ResNet-101} & Teacher-Baseline &  71.0  & 67.1 \\
 & & Zhang et al. \cite{zhang2021improve} & 72.5  & 67.8 \\
 & & Ours (AID) &\textbf{72.6} & \textbf{68.1} \\
\hline
\end{tabular}\\
\label{tab:COCO}
\end{table*}

In addition to mAP performance, we also compared different architectures' efficiency in terms of FLOPs and parameters. The results are shown in Table \ref{tab:efficiency}. It can be observed that our distilled models with the smaller ResNet-50 backbone are more efficient than the corresponding teacher baselines with ResNet-101 backbones. 
In addition to the previously mentioned higher mAPs, our distillation model enjoys an average of 34.4\% reduction in number of parameters and an average of 20.4\% savings in FLOPs.

\begin{table}[]
\footnotesize
\caption{Model complexity (with 224$\times$224 input resolution)}
\centering
\begin{tabular}{c|c|cc}
\hline
\multicolumn{1}{c}{Model} & \multicolumn{1}{c}{Backbone} & Parames(M)  & GFLOPs \\
\hline
\multirow{2}{*}{RetinaNet+}  & ResNet-50 & 32.06 & 10.07   \\
 & ResNet-101 & 51.05 & 13.79   \\
 \hline
\multirow{2}{*}{Faster R-CNN} & ResNet-50 & 41.17 &  23.40   \\
 & ResNet-101 & 60.17 &  27.13   \\
\hline
\end{tabular}
\label{tab:efficiency}
\end{table}

\subsection{Qualitative Analysis}
\input{fig/kitti}
\input{fig/coco_traffic}
Fig. \ref{fig:kitti} shows random qualitative results of our AID self-distilled GFL model and the two other GFL models with ResNet-101 as the backbone on the KITTI dataset. From top to bottom, the results are respectively generated by 1) the teacher baseline GFL model, 2) Zhang \etal~\cite{zhang2021improve}'s model, and 3) our AID-distilled model. For the readers' convenience, we highlighted the prediction differences between our AID-based model and the other baseline models using green ovals. In the upper left image, the teacher baseline model incorrectly detected the fence as a car in the left part of the image. In the upper right image, the teacher baseline model generated a bunch of approximate bounding boxes to locate the pedestrian and the car that overlap each other in the right part of the image. According to the middle two images, Zhang \etal~\cite{zhang2021improve}'s attention-guided method's distilled GFL model performed better than the non-distilled baseline GFL model. In the middle left image, the distilled baseline model (Zhang \etal~\cite{zhang2021improve}) did not make the same mistakes as the teacher baseline model. However, in the middle right image, the model derived by Zhang \etal~\cite{zhang2021improve}'s method still struggles to find the correct bounding boxes for the overlapping objects. Our AID-distilled model not only inherits the benefits of the existing KD approach but also has better detection capability for overlapping objects and small-scale objects. For example, in the down-left image, our AID distilled model did not make the same wrong prediction as the teacher baseline model and had higher confidence scores for small-scale objects at the far end of the street (marked with the smaller green oval). Moreover, in the down-right image, our model correctly detected the pedestrian and the car without generating any false positive prediction. In the same image, our AID-based model also successfully detected the small object behind the pole (as shown in the left green oval), something the previous two competing models failed to achieve.

Fig. \ref{fig:coco traffic} demonstrates another random example on the COCO traffic dataset. From left to right, the results are respectively from 1) the non-distilled student baseline GFL model, 2) Zhang \etal~\cite{zhang2021improve}'s distilled GFL model, and 3) our AID-distilled GFL model. All three models use ResNet-50 as the backbone for fair comparison. The two left images show that 1) and 2) incorrectly predicted the truck in the marked green ovals. The main reason for the detection failure is that the object is occluded, and the teacher model cannot impart trustworthy information to the student model in such scenarios. Zhang \etal \cite{zhang2021improve}'s distilled model blindly trusted the teacher's prediction and thus made a similar mistake. In contrast, our AID-distilled model relied more on itself when learned to predict for such instances and predicted correctly in the rightmost picture.

\section{Future work}

In this paper, we have shown that by adjusting the weight of an instance according to its difficulty in the eyes of the teacher, our AID method can guide the student model to pay more attention to those instances where the teacher does well, enhancing the transfer of reliable and valuable knowledge.
In addition to the baselines that we have experimented with, we believe that our AID approach can be generally applicable to other KD methods (e.g., \cite{wang2019distilling, 8100259}) and detection models (e.g., YOLOs \cite{redmon2018yolov3} and SSD \cite{2016}), and its generalization performance will be tested in our future experiments. We also plan to adopt even smaller backbones, such as ResNet18 and those obtained from pruning approaches~\cite{tian2021task}.
Although some promising results have been achieved by our instance adaptive self-distillation method, we plan to make this process iterative. With more iterations, we expect that a detection model can be further refined.
Last but not least, our AID can potentially help the student selectively learn from multiple teachers, each excels in a particular area (a certain set of instances).

\section{Conclusion}

In this paper, we have proposed an adaptive instance distillation (AID) method to guide the student model to better learn from the teacher model. The method enables the student network to pay more attention to instances that the teacher model performs well on. By applying our AID to different scales of the FPN, we can also make the knowledge distillation process scale-aware.  
Our instance/scale weighting method can potentially be combined with most if not all knowledge distillation methods while inheriting their advantages. In our experiment, we chose the state-of-the-art knowledge distillation approach (Zhang \etal~\cite{zhang2021improve}) as the baseline and tested our AID on both single-stage and two-stage detectors. Experimental results on the KITTI and COCO traffic datasets demonstrate our AID method's efficacy. On average, 2.7\% and 2.1\% mAP increases can be achieved for the single-stage detectors and two-stage detectors, respectively.

\section*{Acknowledgment}
This research was supported by the National Science Foundation (NSF) under Award No. 2153404. This work would not have been possible without the computing resources provided by the Ohio Supercomputer Center.

% trigger a \newpage just before the given reference
% number - used to balance the columns on the last page
% adjust value as needed - may need to be readjusted if
% the document is modified later
%\IEEEtriggeratref{8}
% The "triggered" command can be changed if desired:
%\IEEEtriggercmd{\enlargethispage{-5in}}

%\clearpage
% references section

% can use a bibliography generated by BibTeX as a .bbl file
% BibTeX documentation can be easily obtained at:
% http://mirror.ctan.org/biblio/bibtex/contrib/doc/
% The IEEEtran BibTeX style support page is at:
% http://www.michaelshell.org/tex/ieeetran/bibtex/
\bibliographystyle{IEEEtran}
% argument is your BibTeX string definitions and bibliography database(s)
\bibliography{IEEEabrv,main}
%
% <OR> manually copy in the resultant .bbl file
% set second argument of \begin to the number of references
% (used to reserve space for the reference number labels box)

% \begin{thebibliography}{1}

% \bibitem{IEEEhowto:kopka}
% H.~Kopka and P.~W. Daly, \emph{A Guide to \LaTeX}, 3rd~ed.\hskip 1em plus
%  0.5em minus 0.4em\relax Harlow, England: Addison-Wesley, 1999.

% \end{thebibliography}

\appendix
\section{Appendix}
To ensure the reproducibility of our work, we provide the random train/validation split of the KITTI dataset that we used in our experiments (the test data of KITTI has no ground truth annotations). The validation IDs are (other IDs form the training set):
[\input{fig/val_info}]

The conference \cite{aid2022lan} and journal versions \cite{lan2022aid} of this preprint use the same train/validation split.

% Since the KITTI website does not directly release the ground truth (bounding boxes and labels) of the test dataset, we split those data that have ground truth into training and validation sets. To facilitate others to reproduce our works, we suggest using the following list as the validation set (the remaining data as the training). The conference and the journal versions \cite{lan2022aid} paper use the same train/validation split.

% The Validation List: [\input{fig/val_info}]

% that's all folks
\end{document}

%% file: fig/AID.tex
\begin{figure*}
\begin{center}
% \begin{overpic} 
% [width=\linewidth]
% {example-image-a}
% \end{overpic}
\includegraphics[width=0.7\linewidth]{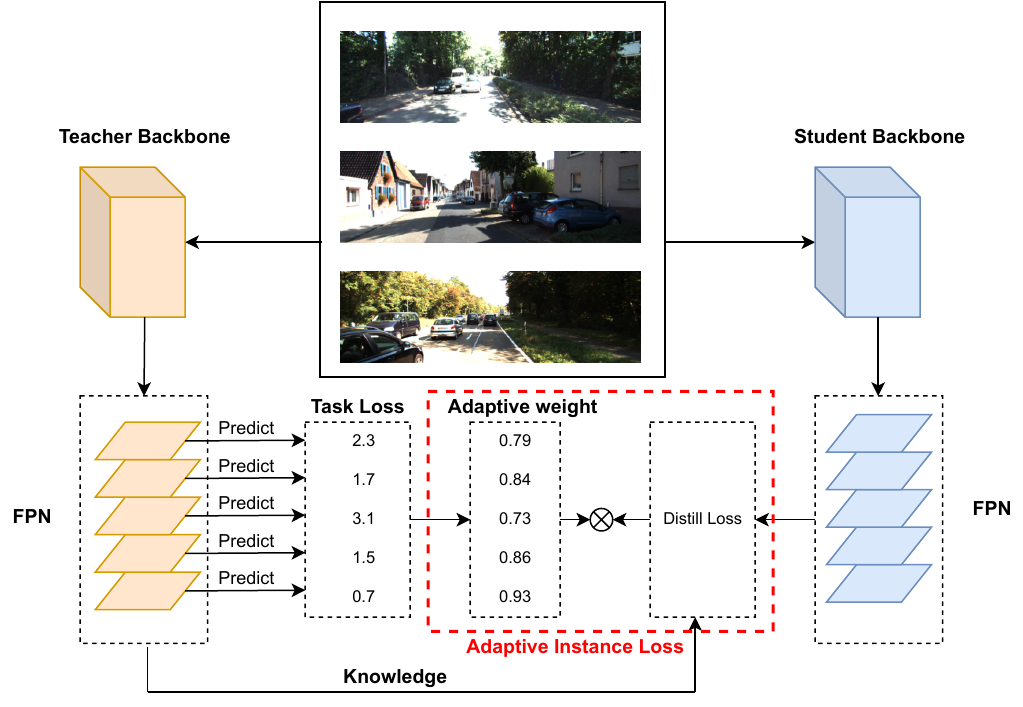}
\end{center}
\caption{
Illustration of the proposed adaptive instance distillation (AID) method. The losses associated with the multi-scale prediction of the teacher model will be transformed into weights to guide the knowledge distillation process.
}
\label{fig:AID}
\end{figure*}

%% file: fig/kitti.tex
\begin{figure*}[]
\begin{center}
\includegraphics[width=.99\columnwidth, trim={0 0 3cm 0},clip]{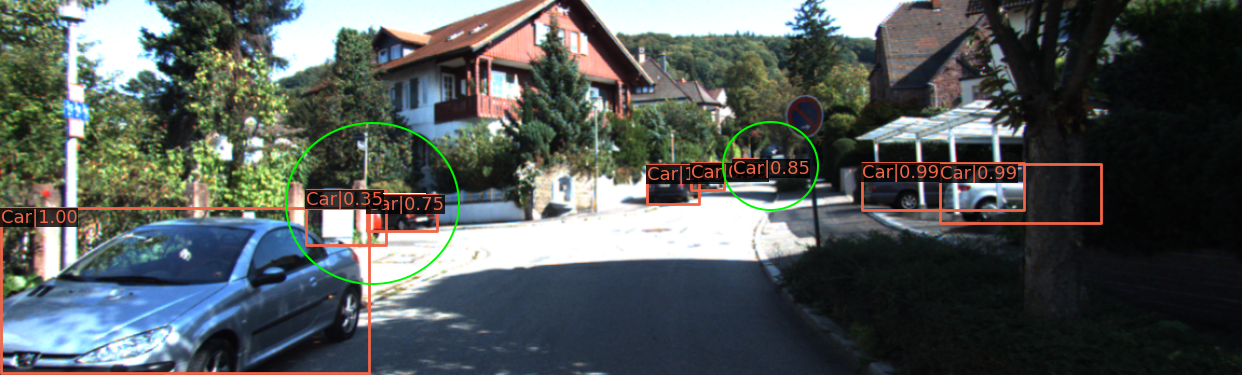}
\includegraphics[width=.99\columnwidth, trim={11cm 1cm 0 1.5cm},clip]{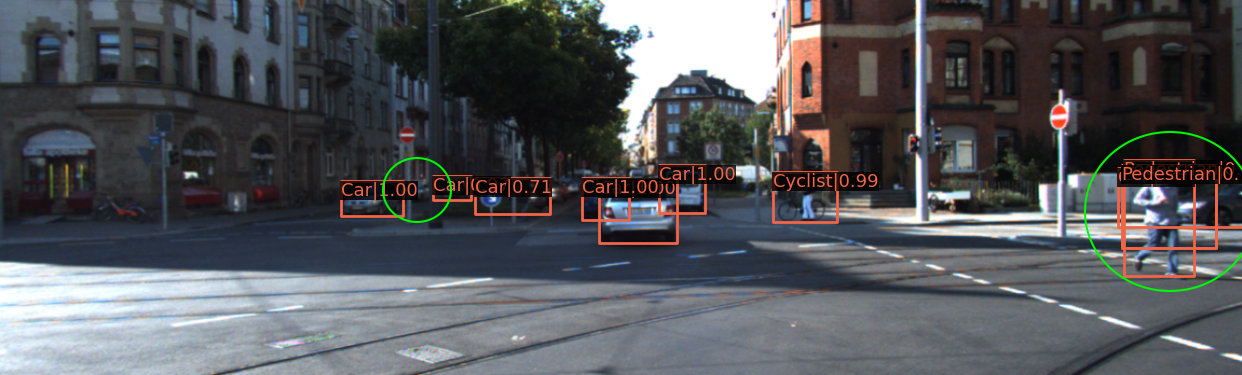}
\hfill\\
\includegraphics[width=.99\columnwidth, trim={0 0 3cm 0},clip]{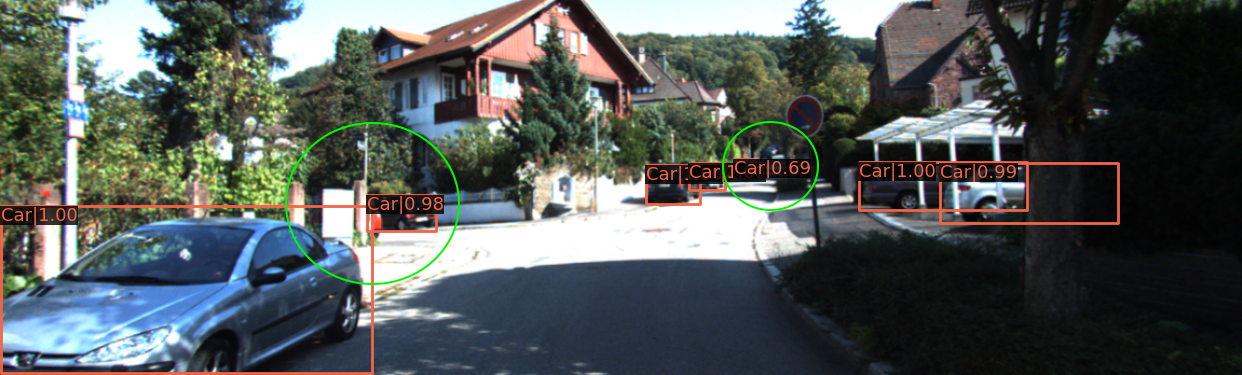}
\includegraphics[width=.99\columnwidth, trim={11cm 1cm 0 1.5cm},clip]{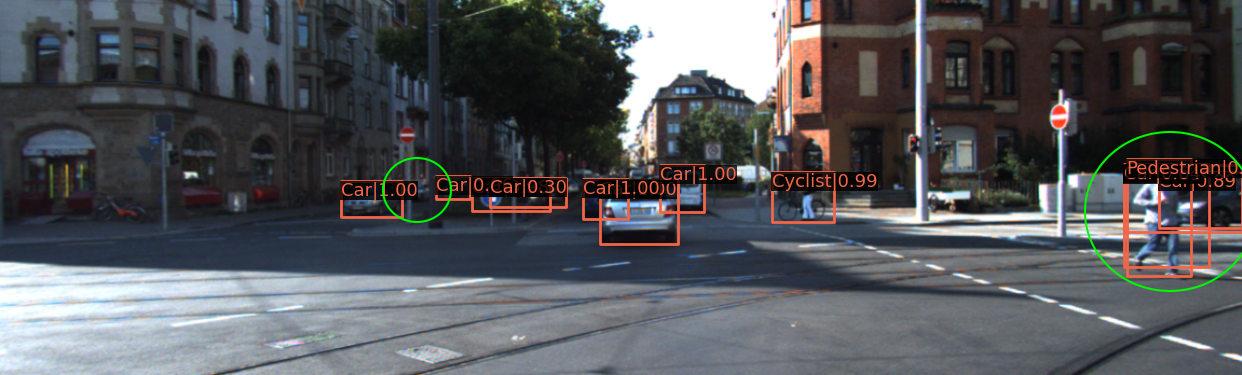}
\hfill\\
\includegraphics[width=.99\columnwidth, trim={0 0 3cm 0},clip]{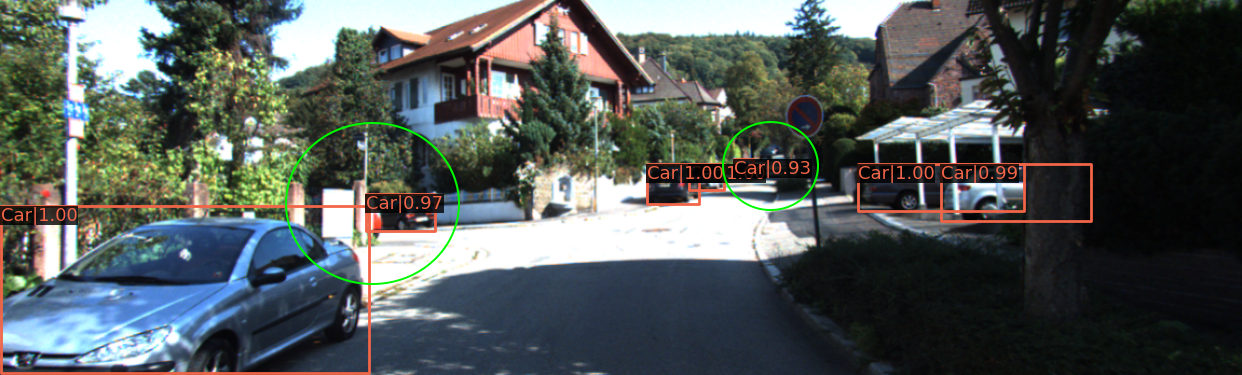}
\includegraphics[width=.99\columnwidth, trim={11cm 1cm 0 1.5cm},clip]{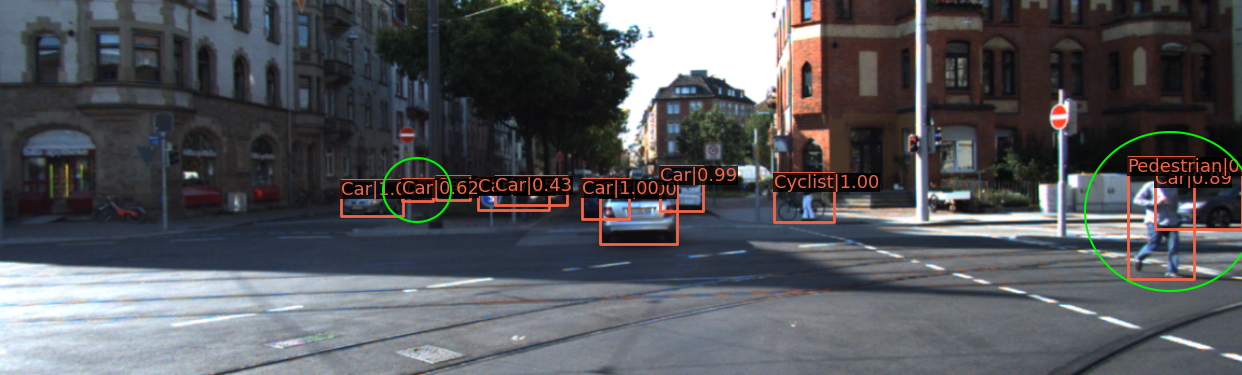}
\end{center}
\caption{
\textbf{Qualitative Analysis on KITTI (best viewed
when zoomed in) -- }
From top to bottom, the prediction results are respectively from 1) Teacher baseline model, 2) Zhang \etal~\cite{zhang2021improve}'s KD baseline model, and 3) our AID distilled model. We use green ovals to highlight some of the detection differences between our AID-based model and other baselines.}
\label{fig:kitti}
\end{figure*}

%% file: fig/coco_traffic.tex
\begin{figure*}[!htp]
\begin{center}
\includegraphics[width=.65\columnwidth, trim={0 2cm 0 2cm},clip]{fig/gfl_000000147725.jpg}
\hfill
\includegraphics[width=.65\columnwidth, trim={0 2cm 0 2cm},clip]{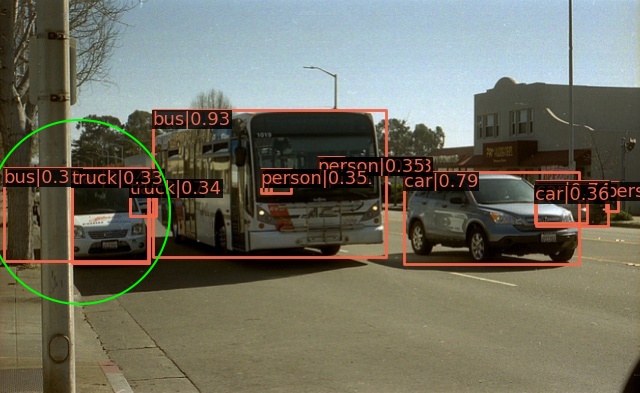}
\hfill
\includegraphics[width=.65\columnwidth, trim={0 2cm 0 2cm},clip]{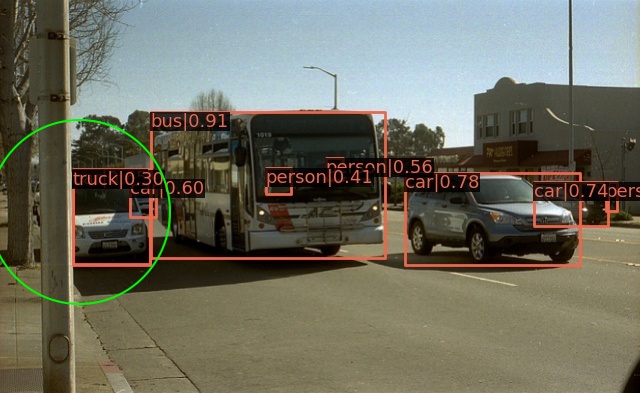}
\end{center}
\caption{
\textbf{Qualitative Analysis on COCO traffic (best viewed
when zoomed in) -- }
From left to right, the prediction results are respectively from 1) Student baseline model, 2) Zhang \etal~\cite{zhang2021improve}'s KD baseline model, and 3) our AID distilled model. We use green ovals to highlight some of the detection differences between our AID-based model and other baselines.}
\label{fig:coco traffic}
\end{figure*}

%% file: fig/val_info.tex
000010
000022
000026
000061
000063
000067
000082
000083
000096
000105
000119
000124
000128
000133
000136
000144
000165
000204
000225
000235
000254
000276
000290
000291
000298
000333
000351
000359
000369
000372
000381
000389
000417
000431
000450
000451
000457
000474
000484
000485
000511
000521
000529
000547
000549
000567
000569
000574
000577
000587
000591
000592
000594
000612
000630
000631
000662
000663
000672
000675
000684
000686
000691
000693
000694
000709
000713
000738
000744
000753
000754
000756
000762
000763
000778
000779
000790
000810
000826
000836
000843
000846
000849
000856
000863
000870
000886
000889
000896
000904
000921
000929
000933
000935
000959
000962
000964
000965
000975
000982
000994
000995
000996
000999
001017
001018
001033
001035
001048
001070
001084
001091
001097
001099
001101
001114
001124
001127
001169
001188
001205
001215
001234
001238
001251
001262
001270
001313
001316
001335
001339
001342
001347
001352
001357
001361
001371
001375
001382
001418
001426
001439
001442
001455
001462
001471
001484
001488
001502
001521
001531
001536
001537
001545
001549
001553
001583
001602
001612
001615
001617
001640
001656
001666
001672
001683
001687
001689
001692
001698
001704
001711
001712
001717
001723
001727
001735
001740
001754
001757
001758
001759
001766
001781
001793
001795
001800
001801
001812
001828
001841
001849
001851
001861
001882
001885
001891
001895
001900
001911
001912
001923
001941
001949
001950
001951
001955
001965
001967
001971
001974
001981
001983
002003
002017
002018
002029
002037
002041
002047
002061
002072
002094
002100
002105
002124
002126
002134
002152
002155
002160
002167
002173
002224
002228
002232
002233
002236
002243
002279
002289
002295
002307
002313
002315
002328
002359
002371
002379
002406
002420
002428
002430
002432
002436
002447
002456
002459
002460
002468
002477
002485
002520
002528
002544
002558
002563
002594
002608
002615
002622
002637
002639
002641
002649
002650
002660
002695
002704
002705
002719
002720
002731
002735
002743
002747
002755
002781
002782
002785
002788
002804
002845
002855
002884
002885
002897
002898
002901
002905
002914
002918
002922
002928
002933
002935
002941
002954
002965
002971
003002
003025
003037
003045
003050
003056
003068
003095
003121
003187
003217
003218
003238
003241
003260
003267
003295
003323
003338
003376
003398
003409
003412
003418
003425
003435
003438
003440
003447
003455
003456
003464
003465
003468
003504
003534
003536
003537
003551
003557
003561
003581
003583
003599
003620
003621
003660
003671
003675
003676
003704
003706
003709
003712
003716
003719
003740
003745
003746
003751
003754
003764
003769
003783
003808
003823
003846
003852
003877
003882
003894
003900
003906
003909
003927
003932
003947
003952
003971
003974
003986
004003
004008
004016
004026
004034
004039
004045
004048
004055
004063
004064
004071
004082
004096
004132
004149
004153
004187
004189
004198
004199
004211
004214
004224
004237
004289
004293
004305
004307
004312
004313
004314
004322
004329
004341
004390
004394
004395
004396
004431
004436
004444
004450
004451
004458
004498
004500
004511
004514
004517
004528
004543
004548
004552
004555
004559
004561
004577
004580
004581
004589
004596
004601
004603
004606
004613
004614
004625
004631
004634
004645
004666
004676
004680
004697
004703
004712
004716
004726
004729
004739
004745
004746
004747
004750
004751
004752
004759
004768
004786
004792
004793
004794
004808
004814
004827
004830
004831
004834
004842
004859
004869
004871
004879
004884
004894
004902
004909
004945
004953
004964
004983
004989
004993
005006
005018
005027
005028
005033
005034
005049
005061
005064
005075
005080
005081
005084
005112
005130
005136
005142
005147
005153
005157
005158
005162
005231
005233
005237
005251
005278
005284
005300
005306
005321
005324
005327
005329
005330
005341
005342
005403
005404
005437
005449
005461
005462
005467
005469
005477
005485
005518
005536
005539
005541
005543
005561
005566
005568
005577
005578
005595
005601
005619
005627
005637
005644
005658
005659
005662
005665
005668
005670
005680
005693
005694
005696
005704
005717
005722
005743
005747
005750
005753
005762
005769
005771
005791
005797
005812
005829
005852
005868
005873
005886
005939
005959
005966
005967
005970
005976
005981
005992
006002
006012
006025
006028
006034
006035
006043
006044
006050
006052
006056
006059
006075
006080
006096
006105
006106
006133
006151
006166
006179
006198
006200
006209
006219
006229
006230
006233
006238
006260
006272
006283
006295
006319
006324
006327
006331
006347
006350
006366
006376
006396
006402
006404
006423
006430
006446
006461
006468
006471
006489
006492
006493
006504
006509
006521
006540
006575
006587
006592
006604
006613
006623
006628
006629
006634
006640
006641
006650
006658
006665
006674
006677
006679
006696
006707
006730
006742
006743
006756
006779
006786
006787
006799
006811
006818
006822
006826
006836
006837
006839
006848
006852
006874
006879
006885
006895
006901
006911
006931
006971
006977
006991
007013
007014
007016
007019
007080
007088
007092
007097
007110
007126
007136
007138
007139
007144
007157
007165
007184
007190
007194
007213
007235
007247
007255
007270
007286
007302
007365
007373
007381
007389
007391
007393
007403
007408
007414
007423
007433
007440
007450
007454
007468

%% file: main.bbl
% Generated by IEEEtran.bst, version: 1.12 (2007/01/11)
\begin{thebibliography}{10}
\providecommand{\url}[1]{#1}
\csname url@samestyle\endcsname
\providecommand{\newblock}{\relax}
\providecommand{\bibinfo}[2]{#2}
\providecommand{\BIBentrySTDinterwordspacing}{\spaceskip=0pt\relax}
\providecommand{\BIBentryALTinterwordstretchfactor}{4}
\providecommand{\BIBentryALTinterwordspacing}{\spaceskip=\fontdimen2\font plus
\BIBentryALTinterwordstretchfactor\fontdimen3\font minus
  \fontdimen4\font\relax}
\providecommand{\BIBforeignlanguage}[2]{{%
\expandafter\ifx\csname l@#1\endcsname\relax
\typeout{** WARNING: IEEEtran.bst: No hyphenation pattern has been}%
\typeout{** loaded for the language `#1'. Using the pattern for}%
\typeout{** the default language instead.}%
\else
\language=\csname l@#1\endcsname
\fi
#2}}
\providecommand{\BIBdecl}{\relax}
\BIBdecl

\bibitem{hinton2015distilling}
G.~Hinton, O.~Vinyals, and J.~Dean, ``Distilling the knowledge in a neural
  network,'' 2015.

\bibitem{krizhevsky2009learning}
\BIBentryALTinterwordspacing
A.~Krizhevsky, ``Learning multiple layers of features from tiny images,'' pp.
  32--33, 2009. [Online]. Available:
  \url{https://www.cs.toronto.edu/~kriz/learning-features-2009-TR.pdf}
\BIBentrySTDinterwordspacing

\bibitem{DBLP:journals/corr/abs-1906-03609}
\BIBentryALTinterwordspacing
T.~Wang, L.~Yuan, X.~Zhang, and J.~Feng, ``Distilling object detectors with
  fine-grained feature imitation,'' \emph{CoRR}, vol. abs/1906.03609, 2019.
  [Online]. Available: \url{http://arxiv.org/abs/1906.03609}
\BIBentrySTDinterwordspacing

\bibitem{DBLP:journals/corr/abs-2102-12252}
\BIBentryALTinterwordspacing
Z.~Zheng, R.~Ye, P.~Wang, J.~Wang, D.~Ren, and W.~Zuo, ``Localization
  distillation for object detection,'' \emph{CoRR}, vol. abs/2102.12252, 2021.
  [Online]. Available: \url{https://arxiv.org/abs/2102.12252}
\BIBentrySTDinterwordspacing

\bibitem{zhang2021improve}
\BIBentryALTinterwordspacing
L.~Zhang and K.~Ma, ``Improve object detection with feature-based knowledge
  distillation: Towards accurate and efficient detectors,'' in
  \emph{International Conference on Learning Representations}, 2021. [Online].
  Available: \url{https://openreview.net/forum?id=uKhGRvM8QNH}
\BIBentrySTDinterwordspacing

\bibitem{DBLP:journals/corr/abs-2103-02340}
\BIBentryALTinterwordspacing
X.~Dai, Z.~Jiang, Z.~Wu, Y.~Bao, Z.~Wang, S.~Liu, and E.~Zhou, ``General
  instance distillation for object detection,'' \emph{CoRR}, vol.
  abs/2103.02340, 2021. [Online]. Available:
  \url{https://arxiv.org/abs/2103.02340}
\BIBentrySTDinterwordspacing

\bibitem{FL}
\BIBentryALTinterwordspacing
T.~Lin, P.~Goyal, R.~B. Girshick, K.~He, and P.~Doll{\'{a}}r, ``Focal loss for
  dense object detection,'' \emph{CoRR}, vol. abs/1708.02002, 2017. [Online].
  Available: \url{http://arxiv.org/abs/1708.02002}
\BIBentrySTDinterwordspacing

\bibitem{li2018gradient}
B.~Li, Y.~Liu, and X.~Wang, ``Gradient harmonized single-stage detector,''
  2018.

\bibitem{cao2019prime}
Y.~Cao, K.~Chen, C.~C. Loy, and D.~Lin, ``Prime sample attention in object
  detection,'' 2019.

\bibitem{zhang2020primeaware}
Y.~Zhang, Z.~Lan, Y.~Dai, F.~Zeng, Y.~Bai, J.~Chang, and Y.~Wei, ``Prime-aware
  adaptive distillation,'' in \emph{Computer Vision -- ECCV 2020}, A.~Vedaldi,
  H.~Bischof, T.~Brox, and J.-M. Frahm, Eds.\hskip 1em plus 0.5em minus
  0.4em\relax Cham: Springer International Publishing, 2020, pp. 658--674.

\bibitem{fpn}
\BIBentryALTinterwordspacing
T.~Lin, P.~Doll{\'{a}}r, R.~B. Girshick, K.~He, B.~Hariharan, and S.~J.
  Belongie, ``Feature pyramid networks for object detection,'' \emph{CoRR},
  vol. abs/1612.03144, 2016. [Online]. Available:
  \url{http://arxiv.org/abs/1612.03144}
\BIBentrySTDinterwordspacing

\bibitem{Vaillant94anoriginal}
R.~Vaillant, C.~Monrocq, and Y.~L. Cun, ``An original approach for the
  localization of objects in images,'' 1994.

\bibitem{990517}
P.~Viola and M.~Jones, ``Rapid object detection using a boosted cascade of
  simple features,'' in \emph{Proceedings of the 2001 IEEE Computer Society
  Conference on Computer Vision and Pattern Recognition. CVPR 2001}, vol.~1,
  2001, pp. I--I.

\bibitem{DBLP:journals/corr/RenHG015}
\BIBentryALTinterwordspacing
S.~Ren, K.~He, R.~B. Girshick, and J.~Sun, ``Faster {R-CNN:} towards real-time
  object detection with region proposal networks,'' \emph{CoRR}, vol.
  abs/1506.01497, 2015. [Online]. Available:
  \url{http://arxiv.org/abs/1506.01497}
\BIBentrySTDinterwordspacing

\bibitem{YOLOV1}
\BIBentryALTinterwordspacing
J.~Redmon, S.~K. Divvala, R.~B. Girshick, and A.~Farhadi, ``You only look once:
  Unified, real-time object detection,'' \emph{CoRR}, vol. abs/1506.02640,
  2015. [Online]. Available: \url{http://arxiv.org/abs/1506.02640}
\BIBentrySTDinterwordspacing

\bibitem{li2020generalized}
X.~Li, W.~Wang, L.~Wu, S.~Chen, X.~Hu, J.~Li, J.~Tang, and J.~Yang,
  ``Generalized focal loss: Learning qualified and distributed bounding boxes
  for dense object detection,'' \emph{arXiv preprint arXiv:2006.04388}, 2020.

\bibitem{DBLP:journals/corr/RedmonF16}
\BIBentryALTinterwordspacing
J.~Redmon and A.~Farhadi, ``{YOLO9000:} better, faster, stronger,''
  \emph{CoRR}, vol. abs/1612.08242, 2016. [Online]. Available:
  \url{http://arxiv.org/abs/1612.08242}
\BIBentrySTDinterwordspacing

\bibitem{yolox}
Z.~Ge, S.~Liu, F.~Wang, Z.~Li, and J.~Sun, ``{YOLOX:} exceeding {YOLO} series
  in 2021,'' \emph{CoRR}, vol. abs/2107.08430, 2021.

\bibitem{FCOS}
\BIBentryALTinterwordspacing
Z.~Tian, C.~Shen, H.~Chen, and T.~He, ``{FCOS:} fully convolutional one-stage
  object detection,'' \emph{CoRR}, vol. abs/1904.01355, 2019. [Online].
  Available: \url{http://arxiv.org/abs/1904.01355}
\BIBentrySTDinterwordspacing

\bibitem{DBLP:journals/corr/abs-2006-02334}
\BIBentryALTinterwordspacing
S.~Qiao, L.~Chen, and A.~L. Yuille, ``Detectors: Detecting objects with
  recursive feature pyramid and switchable atrous convolution,'' \emph{CoRR},
  vol. abs/2006.02334, 2020. [Online]. Available:
  \url{https://arxiv.org/abs/2006.02334}
\BIBentrySTDinterwordspacing

\bibitem{Romero15fitnets:hints}
A.~Romero, S.~E. Kahou, P.~Montréal, Y.~Bengio, U.~D. Montréal, A.~Romero,
  N.~Ballas, S.~E. Kahou, A.~Chassang, C.~Gatta, and Y.~Bengio, ``Fitnets:
  Hints for thin deep nets,'' in \emph{in International Conference on Learning
  Representations (ICLR}, 2015.

\bibitem{huang2017like}
Z.~Huang and N.~Wang, ``Like what you like: Knowledge distill via neuron
  selectivity transfer,'' 2017.

\bibitem{VID2019}
\BIBentryALTinterwordspacing
S.~Ahn, S.~X. Hu, A.~C. Damianou, N.~D. Lawrence, and Z.~Dai, ``Variational
  information distillation for knowledge transfer,'' \emph{CoRR}, vol.
  abs/1904.05835, 2019. [Online]. Available:
  \url{http://arxiv.org/abs/1904.05835}
\BIBentrySTDinterwordspacing

\bibitem{Byeongho2018}
\BIBentryALTinterwordspacing
B.~Heo, M.~Lee, S.~Yun, and J.~Y. Choi, ``Knowledge transfer via distillation
  of activation boundaries formed by hidden neurons,'' \emph{CoRR}, vol.
  abs/1811.03233, 2018. [Online]. Available:
  \url{http://arxiv.org/abs/1811.03233}
\BIBentrySTDinterwordspacing

\bibitem{8100237}
J.~Yim, D.~Joo, J.~Bae, and J.~Kim, ``A gift from knowledge distillation: Fast
  optimization, network minimization and transfer learning,'' in \emph{2017
  IEEE Conference on Computer Vision and Pattern Recognition (CVPR)}, 2017, pp.
  7130--7138.

\bibitem{8953802}
Y.~Liu, J.~Cao, B.~Li, C.~Yuan, W.~Hu, Y.~Li, and Y.~Duan, ``Knowledge
  distillation via instance relationship graph,'' in \emph{2019 IEEE/CVF
  Conference on Computer Vision and Pattern Recognition (CVPR)}, 2019, pp.
  7089--7097.

\bibitem{DBLP:journals/corr/abs-1904-05068}
\BIBentryALTinterwordspacing
W.~Park, D.~Kim, Y.~Lu, and M.~Cho, ``Relational knowledge distillation,''
  \emph{CoRR}, vol. abs/1904.05068, 2019. [Online]. Available:
  \url{http://arxiv.org/abs/1904.05068}
\BIBentrySTDinterwordspacing

\bibitem{Chen2017LearningEO}
G.~Chen, W.~Choi, X.~Yu, T.~X. Han, and M.~Chandraker, ``Learning efficient
  object detection models with knowledge distillation,'' in \emph{NIPS}, 2017.

\bibitem{DBLP:journals/corr/abs-1905-08094}
\BIBentryALTinterwordspacing
L.~Zhang, J.~Song, A.~Gao, J.~Chen, C.~Bao, and K.~Ma, ``Be your own teacher:
  Improve the performance of convolutional neural networks via self
  distillation,'' \emph{CoRR}, vol. abs/1905.08094, 2019. [Online]. Available:
  \url{http://arxiv.org/abs/1905.08094}
\BIBentrySTDinterwordspacing

\bibitem{Geiger2012CVPR}
A.~Geiger, P.~Lenz, and R.~Urtasun, ``Are we ready for autonomous driving? the
  kitti vision benchmark suite,'' in \emph{Conference on Computer Vision and
  Pattern Recognition (CVPR)}, 2012.

\bibitem{lin2015microsoft}
T.-Y. Lin, M.~Maire, S.~Belongie, L.~Bourdev, R.~Girshick, J.~Hays, P.~Perona,
  D.~Ramanan, C.~L. Zitnick, and P.~Dollár, ``Microsoft coco: Common objects
  in context,'' 2015.

\bibitem{mmdetection}
K.~Chen, J.~Wang, J.~Pang, Y.~Cao, Y.~Xiong, X.~Li, S.~Sun, W.~Feng, Z.~Liu,
  J.~Xu, Z.~Zhang, D.~Cheng, C.~Zhu, T.~Cheng, Q.~Zhao, B.~Li, X.~Lu, R.~Zhu,
  Y.~Wu, J.~Dai, J.~Wang, J.~Shi, W.~Ouyang, C.~C. Loy, and D.~Lin,
  ``{MMDetection}: Open mmlab detection toolbox and benchmark,'' \emph{arXiv
  preprint arXiv:1906.07155}, 2019.

\bibitem{wang2019distilling}
T.~Wang, L.~Yuan, X.~Zhang, and J.~Feng, ``Distilling object detectors with
  fine-grained feature imitation,'' 2019.

\bibitem{8100259}
Q.~Li, S.~Jin, and J.~Yan, ``Mimicking very efficient network for object
  detection,'' in \emph{2017 IEEE Conference on Computer Vision and Pattern
  Recognition (CVPR)}, 2017, pp. 7341--7349.

\bibitem{redmon2018yolov3}
J.~Redmon and A.~Farhadi, ``Yolov3: An incremental improvement,'' 2018.

\bibitem{2016}
\BIBentryALTinterwordspacing
W.~Liu, D.~Anguelov, D.~Erhan, C.~Szegedy, S.~Reed, C.-Y. Fu, and A.~C. Berg,
  ``Ssd: Single shot multibox detector,'' \emph{Lecture Notes in Computer
  Science}, p. 21–37, 2016. [Online]. Available:
  \url{http://dx.doi.org/10.1007/978-3-319-46448-0\_2}
\BIBentrySTDinterwordspacing

\bibitem{tian2021task}
Q.~Tian, T.~Arbel, and J.~J. Clark, ``Task dependent deep lda pruning of neural
  networks,'' \emph{Computer Vision and Image Understanding}, vol. 203, p.
  103154, 2021.

\bibitem{aid2022lan}
Q.~Lan and Q.~Tian, ``Adaptive instance distillation for object detection in
  autonomous driving,'' in \emph{2022 26th International Conference on Pattern
  Recognition (ICPR)}, 2022, pp. 4559--4565.

\bibitem{lan2022aid}
------, ``Instance, scale, and teacher adaptive knowledge distillation for
  visual detection in autonomous driving,'' \emph{IEEE Transactions on
  Intelligent Vehicles}, pp. 1--14, 2022.

\end{thebibliography}
